\documentclass[letterpaper, 10 pt, conference]{ieeeconf}  % Comment this line out if you need a4paper

\IEEEoverridecommandlockouts                              % This command is only needed if 
\usepackage{graphics} % for pdf, bitmapped graphics files
\usepackage{epsfig} % for postscript graphics files
\usepackage{amsmath} % assumes amsmath package installed
\usepackage{amssymb}  % assumes amsmath package installed
\usepackage{xcolor}
\usepackage{comment}
\usepackage{multicol}
\usepackage{multirow}
\usepackage{url}
\usepackage{subcaption}
\usepackage{eso-pic} % for overlays
\newcommand{\firstpagecopyright}{%
    \AddToShipoutPictureFG*{%
        \AtPageUpperLeft{%
        \hspace*{\dimexpr1in+\oddsidemargin\relax}%
        \raisebox{-3.5\baselineskip}[0pt][0pt]{%
            \begin{minipage}{\textwidth}
            \centering\footnotesize
            \textit{\textcopyright~2026 IEEE. Personal use of this material is permitted.
            Permission from IEEE must be obtained for all other uses, in any current or future media,
            including reprinting/republishing this material for advertising or promotional purposes,
            creating new collective works, for resale or redistribution to servers or lists, or reuse of
            any copyrighted component of this work in other works.}\\
            Preprint version (Jun.\ 2026). This work has been accepted for publication in IROS 2026.
            \end{minipage}
        }
        }
    }
}

\begin{document}
\firstpagecopyright
\newtheorem{problem}{Problem}
\newtheorem{definition}{Definition}
\newtheorem{example}{Example}
% \newtheorem{problem}{Problem}
% \theoremstyle{definition}
% \theoremstyle{definition}

%%%%%%%%%%%%%%%%%%%%%%%%%%%%%%%%%%%%%%%%%%%%%%%%%%%%%%%%%%%%%%%%%%%%%%%%

%%% Define any new commands you require here.

% \newcommand{\BibTeX}{B\kern-.05em{\sc i\kern-.025em b}\kern-.08em\TeX}
\newcommand{\track}{\tau}
\newcommand{\occupancyPredictor}{\Pi}
\newcommand{\predictorProbability}{\psi}
\newcommand{\occupancyProbabilityFunction}{\omega}
\newcommand{\occupancyProbability}{\rho}
\newcommand{\discreteOccupancyProbabilityFunction}{\bar \omega}
\newcommand{\discreteOccupancyProbability}{\bar \rho}
\newcommand{\occupancyMap}{\Omega}
\newcommand{\discreteOccupancyMap}{\bar \Omega}
\newcommand{\edgeTraverseTimeFunction}{\epsilon}
\newcommand{\discreteEdgeTraverseTimeFunction}{\bar \epsilon}
\newcommand{\visitedVertices}{\nu}
\newcommand{\mdpProbability}{\phi}
\newcommand{\discreteOccupancyLevel}{ \mathcal{\bar L}}
\newcommand{\discreteTimes}{\Delta}
\newcommand{\tour}{\theta}
\newcommand{\tourSet}{\Theta}
\newcommand{\tourCost}{\mathcal{C}}
\newcommand{\personPositionPredictor}{\Psi}
\newcommand{\personPositionPrediction}{\varrho}
\newcommand{\location}{l}
\newcommand{\setOfTracks}{\mathbf{T}}
\newcommand{\setOfPersonPositionPrediction}{\mathbf{P}}
\newcommand{\setOfLocations}{\mathbf{L}}
\newcommand{\addComment}[1]{\textbf{\textcolor{red}{#1}}}
\newcommand{\TODO}[1]{\textcolor{red}{TODO: #1}}
\newcommand{\llbracket}{[\![}
\newcommand{\rrbracket}{]\!]}

% Commands added by Charlie + those he's used from before
\newcommand{\distset}{\text{Dist}}
\newcommand{\edgedist}{\rho}
\newcommand{\topologicalMap}{\mathcal{T}}
\newcommand{\vertices}{\mathcal{V}}
\newcommand{\edges}{\mathcal{E}}
\newcommand{\mdpStates}{S}
\newcommand{\mdpInit}{\bar{s}}
\newcommand{\mdpActions}{A}
\newcommand{\mdpTransitionFunction}{T}
\newcommand{\mdpCost}{C}
\newcommand{\mdpGoal}{G}
\newcommand{\mdp}{\mathcal{M}}
\newcommand{\policy}{\pi}
\newcommand{\timebound}{D}
\newcommand{\wait}{\text{wait}}
\newcommand{\waitdur}{d_{\wait}}
\newcommand{\visitedset}{\omega}
\newcommand{\congprob}{\texttt{cong}(e,j,t)}
\newcommand{\heuristic}{h}
\newcommand{\cliffMapName}{CLiFF map}
\newcommand{\cliffMap}{\Xi}
\newcommand{\cliffLHMPName}{CLiFF-LHMP}
\newcommand{\cliffName}{CLiFF}
\newcommand{\observation}{\mathcal{H}}
\newcommand{\cliffLHMP}{\mathcal{P}}
\newcommand{\predictions}{\Psi}
\newcommand{\rectangle}{R}
\newcommand{\congbandset}{\mathcal{C}}
\newcommand{\band}{c}
\newcommand{\fremen}{FreMEn}
\newcommand{\toCheck}[1]{\textcolor{red}{#1}}
\newcommand{\rewrittenSte}[1]{\textcolor{blue}{ #1 }}

% \keywords{Tour planning, planning under uncertainty, Markov decision processes, congestion, temporal uncertainty.}

%%%%%%%%%%%%%%%%%%%%%%%%%%%%%%%%%%%%%%%%%%%%%%%%%%%%%%%%%%%%%%%%%%%%%%%%

%%% Include any author-defined commands here.
         
% \newcommand{\BibTeX}{\rm B\kern-.05em{\sc i\kern-.025em b}\kern-.08em\TeX}

%%%%%%%%%%%%%%%%%%%%%%%%%%%%%%%%%%%%%%%%%%%%%%%%%%%%%%%%%%%%%%%%%%%%%%%%

\title{Congestion-Aware Robot Tour Planning in Crowded Environments
\author{Stefano Bernagozzi$^{1,2}$, Charlie Street$^{3}$, Masoumeh Mansouri$^{3}$, and Lorenzo Natale$^{2}$}
\thanks{Corresponding Author: {\tt\small  stefano.bernagozzi@iit.it}.\\
This work was funded by the European Union under the Horizon Europe grant 101070227 (CONVINCE).
Charlie Street and Masoumeh Mansouri are UK participants in Horizon Europe Project CONVINCE and supported by UKRI grant number 10042096.
This work was also supported by the Italian National Institute for Insurance against Accidents at Work (INAIL) ErgoCub CORE Project.
%
% For the purpose of open access, the authors have applied a Creative Commons Attribution (CC BY) license to any Accepted Manuscript version arising.
%
\\
$^{1}$Istituto Italiano di Tecnologia -- Genova, Italia
\\
$^{2}$Universit\`a di Genova -- Genova, Italia
\\
$^{3}$University of Birmingham -- Birmingham, United Kingdom
}
% \IEEEauthorblockN{}\\
}
% \fi

%%% The following commands remove the headers in your paper. For final 
%%% papers, these will be inserted during the pagination process.

% \pagestyle{fancy}
% \fancyhead{}

%%% The next command prints the information defined in the preamble.
\maketitle 

\begin{abstract}

Autonomous mobile service robots are often required to complete tours that require navigating through a set of locations in an environment.
Example domains include guiding people through a shopping mall, delivering packages in a fulfilment centre, or giving guided tours in a museum.
However, in crowded environments, the presence of people may negatively impact robot performance.
For example, humans will activate robot collision avoidance maneuvres that slow the robot down.
Crowds move stochastically and vary throughout the day.
In this paper we present a probabilistic tour planner for crowded environments which explicitly reasons over human congestion.
We learn circular linear flow field (CLiFF) maps which predict human trajectories given an initial observation.
We then use these predictions to build and solve a Markov decision process online which efficiently routes the robot through the environment.
Our approach is scalable enough to re-plan as new people are observed.
We evaluate our approach on a real-world crowd dataset in a shopping mall and a physical simulation of a real museum.
\end{abstract}

\section{INTRODUCTION}
\label{sec:introduction}
Autonomous mobile robots have been deployed broadly across the service industry, from airports 
\cite{triebel2016spencer,joosse2017guide} to hotels 
\cite{pan2013direct, choi2020service}.
One particularly prevalent domain has been in \emph{museums} \cite{del2019lindsey,rosa2024tour, al2016tour,velentza2020museum}, where robots provide autonomous guided tours explaining different \emph{points of interest} (POIs).
To facilitate autonomous tours, robots require a \emph{tour planner} that finds efficient routes covering all POIs in the museum, where efficiency is defined in terms of tour duration.
%
%Efficiency here is defined in terms of tour duration; robot tours should have a similar duration to a human-led tour.
%
A major challenge for museum tour planning is the presence of human crowds.
Crowds move stochastically through the environment and vary throughout the day. 
Crowds negatively affect robot navigation as the robot has to use collision avoidance maneuvers to avoid humans, slowing it down.
Therefore, planners should explicitly reason over the effects of crowd congestion to synthesise more efficient tours.
%
%For example, if an artwork is often crowded around midday, the robot should leave that artwork until later in the afternoon.
\begin{example}
% Consider the museum tour guide robot in Fig.~\ref{fig:museum_robot}.
Consider a robot that acts as a tour guide in a museum.
The robot is due to start a tour at $10$am.
If the robot follows the shortest route around the museum, it will reach a popular artwork around $11$am.
However, data shows that this artwork is often crowded around this time.
Navigating to this artwork at $11$am will increase the tour duration and decrease visitor satisfaction.
Therefore, the robot should visit that artwork later in the day when it is less crowded.
\end{example}
In this paper, we present a congestion-aware tour planner that explicitly reasons over human movement to improve tour performance.
We begin by learning a circular linear flow field (CLiFF) map from real human data collected in the environment~\cite{cliff_main}.
Given an initial observation of a human, the CLiFF map predicts their future trajectory.
With this, we present a receding horizon probabilistic tour planner.
At each decision step, we observe the humans in the environment and predict their trajectories using the CLiFF map.
We then build and solve a Markov decision process (MDP)~\cite{puterman2014markov} which uses the CLiFF predictions to reason over which areas will be congested, when they will be congested, and what effect this will have on robot navigation performance.
By using an online, receding horizon planner, our approach is reactive to humans entering and leaving the environment during execution.\looseness=-1
% \begin{figure}
% \centering
%     \includegraphics[scale=0.04]{figures/R1_in_museum.jpg}
%     \caption{A robot tour guide in a museum.}
%     % \description{A robot tour guide in a museum.}
%     \label{fig:museum_robot}
% \end{figure}
The core contribution of this paper is an online framework for tour planning in crowded and congested environments.
Though we focus on guided museum tours as a running example, our approach can be applied for tour planning in any human-populated environment.
We demonstrate the efficacy of our approach on a real-world dataset collected in a shopping mall and a physical simulation of a real museum.

%in the I section we introduce the problem, in the II section we explain the state of the art, in section 3 we explain the algorithm used, section 4 is the comparison with state of the art algorithm and the experimental evaluation and the last section is a brief discussion and the future work based on this paper.
% \cite{fuzzyhumanaware} where the authors have implemented a fuzzy controller to enable human-aware navigation; or \cite{downthecliff} where the authors have exploited a map of dynamics to predict and follow the correct flow direction.
\section{RELATED WORK}\label{sec:related_work}

% \TODO{I think this section needs a lot of work.}
% \TODO{The section does not discuss limitations in the approaches discussed. There isn't much structure to the section. There isn't enough arguing why our method is necessary. There isn't much technical information on how these other approaches work.}
% \TODO{Id recommend splitting this section into 3 paragraphs - one on maps of dynamics and why we choose CLiFF; one on social navigation, highlighting that we're doing higher-level tour planning; and one on higher-level tour planning, and what gap in this literature our method suggests.}

In this section we discuss maps of dynamics that model human motion, social navigation approaches for navigating amongst humans, and task planning methods that explicitly reason over the effects of congestion.

\iffalse
Human-aware robot planning is a widely researched field, and in the latest years there has been many advances in this topic. From social navigation to map of dynamics \cite{kucner2023survey, swaminathan2022benchmarking} the field is increasingly gaining interest in the community. Even though all those fields are closely related, we can distinguish the literature in three main categories: map of dynamics, social navigation and high level tour planning.\\
\fi

%Many approaches focus mainly on few humans and how to predict their behaviour, hence they have a very narrow time window or they focus on their gestures instead of their future positions.

\subsection{Map Of Dynamics}
\label{sub:MoDs}

A map of dynamics captures the dynamic features of an environment that impact a robot, such as human motion for ground robots, wind direction for aerial robots, and currents for underwater robots.
\fremen{}~\cite{krajnik2017fremen} models periodic temporal dynamics using Fourier transforms.
This is suitable for human-populated environments such as offices, where events occur with a regular frequency, such as people leaving for lunch around midday.
However, \fremen{} only models Bernoulli variables such as occupancy, and so is unsuitable for the problems in this paper, where we need to model the number of people in an area at a given time.
This is mitigated in~\cite{vintr2019spatio}, which extends \fremen{} using a warped hypertime representation to admit continuous variables.
\fremen{}~\cite{krajnik2017fremen} and the warped hypertime model~\cite{vintr2019spatio} are only suitable for predictions over longer horizons such as days, weeks etc.
Moreover, they don't explicitly consider motion.
This is partially mitigated in~\cite{molina2018modelling}, where \fremen{} is extended to consider the direction of human motion.
An alternative approach for modelling human motion is to use CLiFF (circular linear flow field) maps, which model the speed and direction of human motion at discrete locations in the environment using semi-wrapped Gaussian mixture models (SWGMMs)~\cite{cliff_main}.
CLiFF maps can be used to predict human trajectories given a history of observations.
In~\cite{cliff-lhmp}, human directions are sampled from the SWGMMs to accurately predict trajectories up to $50$ seconds into the future.
We use CLiFF maps in this paper to predict multiple trajectories for each human and estimate distributions over congestion, i.e. how many humans are in an area at a given time.
Similar flow-based maps of dynamics include~\cite{senanayake2020directional}, which captures the speed and direction of human motion on a directional grid map.
Moreover,~\cite{saraydaryan2023human} predict human trajectories similar to~\cite{cliff-lhmp}, but using von Mises distributions instead of SWGMMs.
Recent work has considered deep learning architectures for modelling crowds.
For example,~\cite{kazemi2025lightweight} use a encoder-forecaster model to predict crowd motion for social navigation.
For further information on maps of dynamics, we refer the reader to a recent survey~\cite{kucner2023survey}.
\subsection{Social Navigation}

Robots deployed in public spaces must safely interact with people with differing levels of exposure to robotic systems.
This poses significant challenges for robot navigation, where the robot must move around people whose intentions may be unclear.
Crowd navigation is often addressed using classical planning~\cite{feyzabadi2014risk, singamaneni2021human, banisetty2021socially} or reinforcement learning-based approaches~\cite{chen2017socially,kastner2022enhancing, wang2022feedback, 8460851}.
Both approaches have been used successfully, however classical approaches are more common due to their low data requirements.
%but the most common ones are still the classical planning approaches because of the low amount of data needed compared to the other ones.
%
% \TODO{Stefano, if you have time, it looks like there was a sentence about classical planning approaches here you never got chance to finish? There's not much description of what different methods do here? If you have time, expanding it a little could be nice}
%
Recent work~\cite{matsumoto2024crowd} combines learning and rule-based methods for crowd navigation, where a high-level planner plans proactively while reinforcement learning is used to avoid local obstacles~\cite{debnath2025hybrid}.
For further information on social navigation, we refer the reader to a number of recent surveys~\cite{singamaneni2024survey, moller2021survey, kruse2013human, charalampous2017recent, mavrogiannis2023core, chik2016review}.

\subsection{Task-Level Planning under Congestion}

Social navigation techniques often focus on handling humans at the motion level.
In this paper, we focus on a higher level of abstraction, where robots make navigation decisions between POIs in a museum based on crowd modelling.
%
% \TODO{Stefano - everything in blue needs some work but I'm finding it hard to edit - it seems there may be missing citations? Also, you say 'the only work that does this' and then seem to discuss other works that do that thing immediately after that sentence?}
%Social navigation techniques often embody robots with social awareness at the motion level. Though we consider social awareness in terms of navigating through crowds, we do this at a higher level of abstraction, where robots make navigation decisions between POIs on a topological map.
%
Few works consider high-level planning in crowded environments.
\cite{dugas2020ian} present a Monte-Carlo tree search-based planner that selects high-level navigation behaviours dependent on the crowdedness of the environment and perceptions of how likely humans are to let the robot through.
Navigation behaviours include standard motion planning, the robot verbally communicating its goal, and nudging forward to escape crowded areas.
This results in the robot navigating through crowds, compared to our approach which navigates around them.
\cite{debnath2025hybrid} present a global and local planner for navigating through human-populated environments.
Similar to our approach, the cost of navigation is increased in areas where humans are observed.
However, this is modelled in an ad-hoc way by applying a hand-designed Gaussian over each observed human location.
In this paper, we model congestion using a \cliffMapName{}~\cite{cliff_main} learned from empirical human data.
%
%To the best of our knowledge, the only works that did a long term high level planning using pedestrian prediction are \cite{dugas2020ian} and \cite{debnath2025hybrid}. In \cite{dugas2020ian} the authors considers the position of the crowd and, differently from our approach, the interaction modalities with it, to make a path through the crowd instead of avoiding it. In \cite{debnath2025hybrid} the authors use a Proactive Global Planner with soft costs to anticipate pedestrian locations, which is similar to our approach, but they use also a local planner to adjust the plan as it is executed. The main difference is that they use a Reinforcement Learning approach that needs to be trained, while our approach uses only a pre-computed \cliffMapName with data of going from one point to the other. 
%Similarly to the approach in \cite{dugas2020ian}, authors in \cite{obo2018intelligent} have implemented a fuzzy controller to enable a multi-objective task planner for a better navigation of the robot in crowded environments, but, differently from us, it does not consider occupancy predictions in their work.
%Other works but from a different field are based on house occupancy for delivery planning like \cite{ohsugidelivery_route} and \cite{teng_optimal_delivery} where the authors have exploited electricity consumption to estimate the time when people are at home and optimize the deliveries and plan the path accordingly. 
%
\cite{congawarestreet} presents an approach for planning under congestion caused by other robots.
Here, robot congestion is modelled using continuous-time Markov chains, with this information being fed into MDPs for planning.
We use a similar MDP-based approach in this paper, where human congestion is modelled using a \cliffMapName{}~\cite{cliff_main}.

\section{PRELIMINARIES}
\label{sec:preliminaries}

In this paper, $\distset(X)$ denotes the set of distributions over $X$.
Further, we write $\llbracket i,j \rrbracket$ to denote the discrete interval $\{i,\cdots,j\}$.

\subsection{Topological Maps}\label{sub:topmap}
In this paper, we use a topological map with distributions over navigation durations to represent the environment for tour planning.

\begin{definition}
\label{def:topmap}
    A \emph{topological map} is a tuple $\topologicalMap = \langle \vertices, \edges, \edgedist \rangle$, where $\vertices$ is a finite set of nodes representing \emph{POIs} in the environment; $\edges \subseteq \vertices \times \vertices$ is a set of edges the robot can navigate along; and $\edgedist: \edges \times \mathbb{N} \rightarrow \distset(\mathbb{R}_{\geq 0})$ is a function that takes an edge and the number of humans on that edge, and returns a distribution over how long it takes the robot to traverse that edge. These distributions can be learned from data.
\end{definition}

\subsection{Markov Decision Processes (MDPs)}\label{sub:mdp}

MDPs model systems with non-deterministic action choices and stochastic action outcomes~\cite{puterman2014markov}.
In this paper, we use MDPs as our model for tour planning in crowded environments.
Tour planning can be formulated using a subset of MDPs known as \emph{stochastic shortest path (SSP)} MDPs.

\begin{definition}
    An \emph{SSP MDP}~\cite{puterman2014markov} is a tuple $\mdp = \langle \mdpStates, \mdpInit, \mdpActions, \mdpTransitionFunction, \mdpCost, \mdpGoal \rangle$, where $\mdpStates$ is a finite set of states; $\mdpInit$ is the initial state; $\mdpActions$ is a finite set of actions; $\mdpTransitionFunction: \mdpStates \times \mdpActions \times \mdpStates \rightarrow [0,1]$ is a probabilistic transition function, where $\mdpTransitionFunction(s,a,s')$ denotes the probability of reaching state $s'$ after executing action $a$ in state $s$; $\mdpCost: \mdpStates \times \mdpActions \rightarrow \mathbb{R}_{\geq 0}$ is a cost function, where $\mdpCost(s,a)$ is the cost of executing action $a$ in state $s$; and $\mdpGoal \subseteq \mdpStates$ is a set of goal states.
\end{definition}

The objective of an SSP MDP is to minimise the expected cumulative cost to reach a goal state.
The optimal solution to an SSP MDP can be represented as a \emph{deterministic memoryless policy}.

\begin{definition}
A \emph{deterministic memoryless policy} is a mapping $\policy: \mdpStates \rightarrow \mdpActions$, where $\policy(s)$ denotes the action to execute in state $s$.
\end{definition}

\subsection{Circular Linear Flow Field (CLiFF) Maps}\label{sub:cliff}

CLiFF maps capture human motion dynamics in terms of their velocity at discrete locations in an environment~\cite{cliff_main}.
In this paper, we use CLiFF maps to predict human congestion during tour planning.

%A CLiFF-map \cite{cliff_main} is a Map of Dynamics (see Section~\ref{sub:MoDs}) that represents for each location the speed and orientation together to indicate the probability of the speed and orientation of a pedestrian at that point. At each location a SemiWrapped Gaussian Mixture Model (SWGMM) represents these data.
\begin{definition}
\label{def:cliff_map}
A \emph{CLiFF map}~\cite{cliff_main} is a set $ \cliffMap = \lbrace \cliffMap_1, \dots, \cliffMap_n \rbrace$, where $\cliffMap_i = \langle l_i, \xi_i, p_i, q_i \rangle.$
Each $\cliffMap_i$ corresponds to a discrete location $l_i \in \mathbb{R}$ in the environment.
Locations $l_i$ are often arranged in a grid.
$\xi_i$ is an SWGMM that describes the distribution over human velocity at $l_i$.
Here, velocity is represented in terms of its orientation (the circumference of the SWGMM) and speed (the height of the SWGMM).
CLiFF maps are data-driven models.
Value $p_i$ describes the probability of motion occurring at $l_i$, i.e. how often was there motion when we observed $l_i$?
Value $q_i$ is the proportion of the data collection time we spent observing $l_i$, and acts as a confidence measure for $\cliffMap_i$.
%A CLiFF-map can be represented as a set $ \cliffMap = \lbrace \cliffMap_1, \dots, \cliffMap_n \rbrace | n \in \mathbb{N} $ and $\cliffMap_i = \langle l_i, \xi_i, p_i, q_i \rangle $ where $l_i \in \mathbb{R}^2 $ is a location on the plane, $\xi$ is a  SWGMM, $p~\in~[0,1]$ is the observation ratio that indicates if the position was observed for the entire duration of the collection and $q~\in~[0,1]$ is the motion ratio that indicates how much motion was observed in that location with respect to the total observation of that location.
\end{definition}

To learn a CLiFF map from continuous data, observations are assigned to any $l_i$ they fall within a fixed radius of.
Given a CLiFF map,~\cite{cliff-lhmp} proposes an approach for predicting human trajectories given a history of observations of their position and velocity.
This history is used to estimate the human's current state, i.e. their position and velocity.
The predicted trajectory is built iteratively along each timestep in the prediction horizon.
For each iteration, they first apply the human's current velocity to obtain their next position.
They then sample a direction and speed from the SWGMM for the nearest location $l_i$ in the CLiFF map.
The human's next direction and speed are then computed by biasing their current direction and speed with those sampled.
%
% The speed is assumed to remain constant.
\begin{definition}
\label{def:cliff-lhmp}
Given a CLiFF map $\cliffMap$ and human observation history $\observation$, we can \emph{predict a trajectory} $\predictions: \mathbb{R}_{\geq0} \rightarrow \mathbb{R}^2$ for the human, where $\predictions(t)$ is the human's predicted location at time $t$.
\end{definition}
\iffalse
Based on the CLiFF-map, \textcite{cliff-lhmp} have implemented a Long Term human-Motion Prediction (LHMP) that given a set of position observation for a person and a CLiFF-map, it outputs a set of predictions that represents the possible future positions of that person.
\begin{definition}
\label{def:cliff-lhmp}
    A CLiFF-LHMP can be represented as a mapping $\cliffLHMP: \langle \observation,\cliffMap \rangle \rightarrow \Psi $ where $\observation$ is a set of person last positions, $\cliffMap$ is a map as Definition \ref{def:cliff_map} and $\predictions \subset \mathbb{R}^2 \times \mathbb{N} \times \mathbb{N} \times [0,1]  $ is a prediction of the future position, where $\predictions(t, n) = x, y, p  $ denotes that the $n^{th}$ predicted position at time $t$ is $\langle x, y \rangle$ with probability $p$ and 
    \begin{equation}
        \sum_{n\in\mathbb{N}} \predictions(t, n) = \begin{cases}
            1 & if \predictions(t, n) \neq \emptyset \\
            0 & if  \predictions(t, n) = \emptyset \\
        \end{cases} \forall t \in \mathbb{N}
    \end{equation}
\end{definition}
\fi
Trajectories $\predictions$ predicted by CLiFF map $\cliffMap$ are sampled from a wider distribution of trajectories for the human.
In Sec.~\ref{sub:congestion_model}, we estimate this distribution by sampling multiple trajectories for each human.
This is used to build a probabilistic model of human congestion.
Further, CLiFF maps discretise the environment in a more granular way than topological maps (see Sec.~\ref{sub:topmap}).
In Sec.~\ref{sub:congestion_model}, we describe how to map predicted trajectories onto topological edges for congestion modelling.

\section{TOUR PLANNING IN CROWDED ENVIRONMENTS}
\label{sec:proposed_algorithm}

In this section, we outline our probabilistic approach for tour planning in crowded environments.
We begin by formally defining the tour planning problem, which requires the robot to reason over edge durations that change stochastically over time due to the effects of human congestion. 
This is similar to a Hamiltonian path problem~\cite{sevaux2008hamiltonian}, but where POIs may be visited more than once.

\begin{problem}\label{prob:tour}
Given a topological map $\topologicalMap$ describing the POIs in the environment, synthesize a policy $\policy$ that minimizes the expected duration to visit each node/POI $v \in \vertices$. POIs may be revisited during the tour.
\end{problem}

\subsection{Solution Overview}

% \begin{figure}
%     \centering
%     \includegraphics[width=0.9\linewidth]{figures/tour_framework.pdf}
%     \caption{A flow chart describing our congestion-aware tour planning framework.}
%     % \Description{A flow chart describing our congestion-aware tour planning framework.}
%     \label{fig:framework}
% \end{figure}

% We present an overview of our tour planner in Fig.~\ref{fig:framework}.
We begin with an overview of our tour planner for crowded environments.
At each decision step, we begin by observing the humans currently in the environment.
This could be achieved using onboard sensors, or external sensors located around the environment.
In this work we consider external sensors that show all humans currently in the environment.
%
%Using onboard sensors, in fact, would reduces the humans that  can detected and limit their position to areas that are quite narrow. Due to the temporal limits of human trajectory prediction combined with online replanning, prediction trajectories solely of humans in the robot field of view, is unlikely to significantly affect tour performance.
%
Given a CliFF map~\cite{cliff_main} computed offline that captures human motion dynamics, we use the prediction mechanism in~\cite{cliff-lhmp} to produce a distribution over trajectories for each observed human.
From the predicted trajectories, we assign each human probabilistically to an edge on the topological map, and then compute a temporal distribution over the number of people on each edge.
This forms our model of congestion.
With this, we incrementally build and solve an MDP for tour planning that uses the congestion distributions to reason over the effect of crowds on robot navigation.
We then execute the first action of the resulting policy before repeating this process at the next decision step.
We proceed by describing our CliFF map-based approach to modelling congestion in Sec.~\ref{sub:congestion_model}, and our MDP planning approach in Sec.~\ref{sub:planning}.

\iffalse
To solve the problem explained in the section above, we have implemented an MPD planner which reasons about the current occupancy, similar to the one defined in \cite{congawarestreet}. Our planner specifically exploit LRTDP to avoid to compute the whole MDP at the beginning, instead it is computed only when certain states are reached.
The planner starts by getting the current tracking of people. In our case the predictor is based on \cite{cliff-lhmp}, where they first build a CLiFF map \cite{cliff_main}, then they sample the possible directions from the given track and finally they return a list of possible tracks. Once we have the tracks, at each time the position is assigned to the edges and after that the Poisson binomial distribution for each edge is returned. 
After that, the algorithm uses those tracking to predict the future positions of the people and it computes the various discrete occupancy maps $\discreteOccupancyMap_t$  various times. 
Using this constructs the states of the MDP dynamically and uses LRTDP to get the best action for the current state. 
Finally it executes the action and at the next vertex re-plan again until the whole map is covered.  

An example of the pipeline of the algorithm can be seen in Figure \ref{fig:framework}.
\fi

\subsection{Modelling Congestion}\label{sub:congestion_model}

\begin{figure}
\centering
    \includegraphics[scale=0.25]{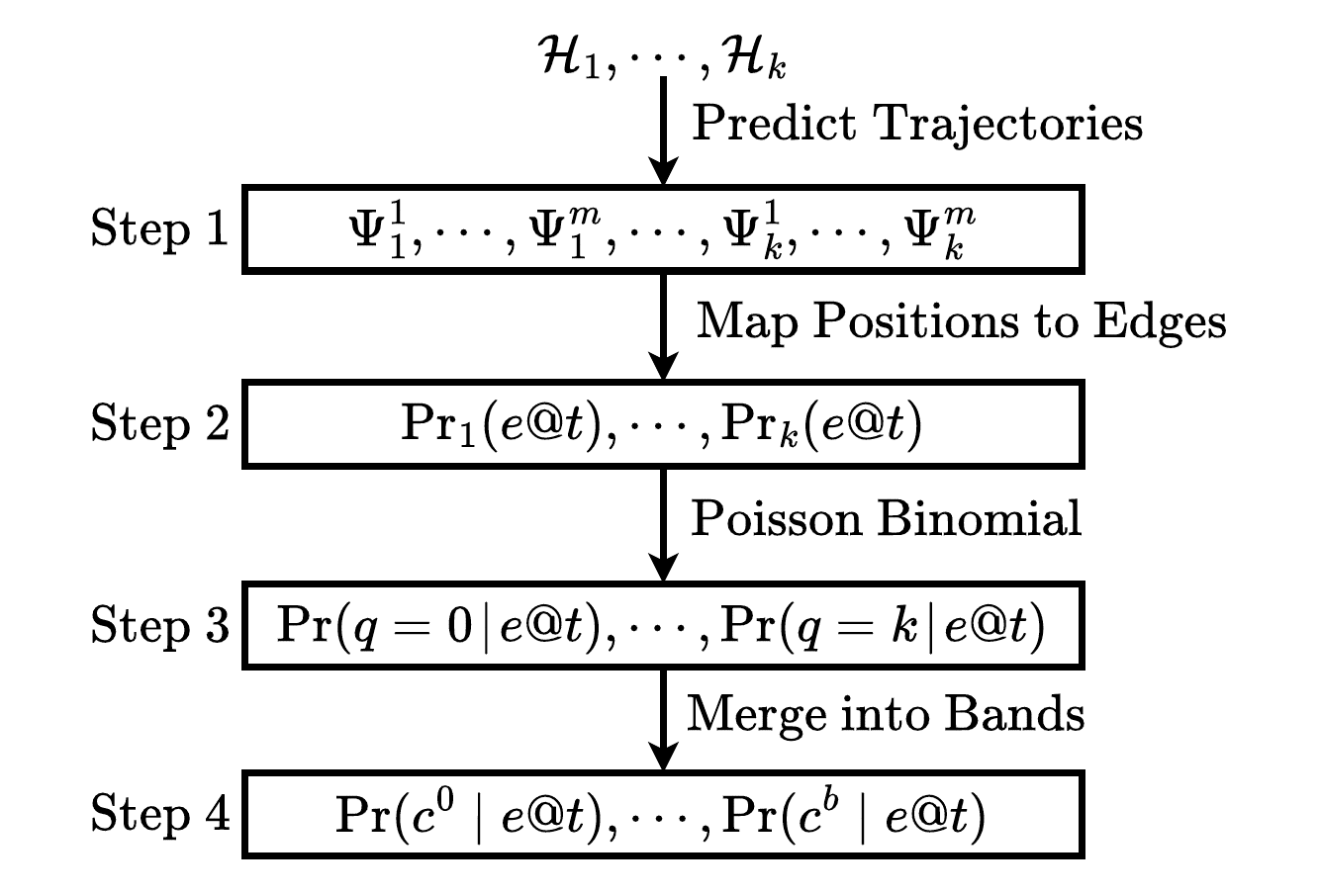}
    \caption{Our approach for computing human congestion probabilities from a CLiFF map.}
    % \description{Our approach for computing human congestion probabilities from a CLiFF map.}
    \label{fig:congestion_model}
 \end{figure}

In this subsection, we describe how to predict and model congestion to support planning.
In particular, we compute the probability of different congestion levels on a topological edge $e$ at time $t$. 
Combining this with the robot duration distributions in Def.~\ref{def:topmap} allows the robot to reason over how long it will take to traverse an edge at time $t$.
We summarize the process for computing congestion probabilities in Fig.~\ref{fig:congestion_model}.

The input to Fig.~\ref{fig:congestion_model} is a set of observation histories $\observation_1,\cdots,\observation_k$ for the $k$ humans currently observed in the environment.
With this, the first step is to predict $m$ trajectories for each human using the CLiFF-LHMP approach outlined in Sec.~\ref{sub:cliff}~\cite{cliff-lhmp}.
For a given human, each trajectory is assigned probability $\frac{1}{m}$.
Human motion is often uncertain and erratic, and so predicting a single trajectory for each human would produce an inaccurate congestion model.
By sampling multiple trajectories, we approximate the underlying trajectory distribution.
This admits a congestion model that explicitly captures motion uncertainty.

In Sec.~\ref{sub:planning} we synthesize tour plans over a topological map.
This is a higher level of abstraction than the trajectories we sample with CLiFF-LHMP, which predict a human's position in 2D space.
Therefore, the second step in computing congestion probabilities is to map predicted human positions onto topological edges.
For this, we build a rectangle $\rectangle_e$ around each edge $e$, where the edge lies in the middle of the rectangle.
A human is on $e$ at time $t$ if their position lies inside $\rectangle_e$.
As rectangles for different edges may overlap, a human may be on multiple edges at once.
This captures the idea that if two edges are close together, a human walking between them may affect robot navigation on either edge.

\begin{figure}
\centering
    \includegraphics[scale=0.4]{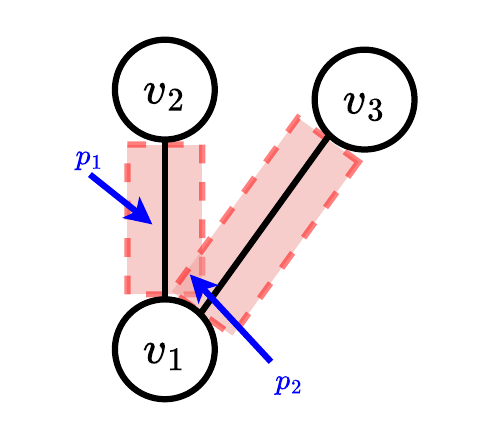}
    \caption{Mapping human positions to topological edges. Position $p_1$ is mapped to $(v_1,v_2)$. Position $p_2$ is mapped to both $(v_1,v_2)$ and $(v_1,v_3)$.}
    \label{fig:edge_mapping}
 \end{figure}

\begin{example}
\label{ex:edge_mapping}
Consider the topological map in Fig.~\ref{fig:edge_mapping}.
The red boxes show the rectangles $\rectangle_e$ expanded around each edge.
Position $p_1$ is on edge $(v_1,v_2)$ as it lies inside $\rectangle_{(v_1,v_2)}$.
Position $p_2$ is on edge $(v_1,v_2)$ and $(v_1,v_3)$, as it lies in the intersection of $\rectangle_{(v_1,v_2)}$ and $\rectangle_{(v_1,v_3)}$.
\end{example}
\noindent Given a mapping from predicted positions to topological edges, we can compute the probability that human $i$ is on edge $e$ at time $t$ by counting the sampled trajectories that fall inside $\rectangle_e$ and multiplying by $\frac{1}{m}$:

\begin{equation}\label{eq:presence_prob}
{\Pr}_i(e@t) = \frac{1}{m} \cdot \sum_{x \in \llbracket1,m\rrbracket} \predictions_i^x(t) \in \rectangle_e.
\end{equation}

Step three in computing the congestion probabilities is to compute the distribution over how many humans are on edge $e$ at time $t$.
This describes how congested the edge will be.
For this we use the Poisson binomial distribution, which models the outcome of multiple trials, where each trial is described using a different Bernoulli random variable~\cite{biscarri2018simple}.
In this instance, the Bernoulli random variables model the probability of each human being present on $e$ at time $t$, as computed by~\eqref{eq:presence_prob}.
We write $\Pr(q\mid e@t)$ to denote the probability of $q$ humans being present on edge $e$ at time $t$.

During execution, there may be little impact on robot navigation duration for similar numbers of humans.
For example, for an edge in a wide corridor, the effect of three humans being present will be similar to four.
Therefore, in the final step of Fig.~\ref{fig:congestion_model}, we merge the probabilities for each number of humans $\Pr(q\mid e@t)$ into \emph{congestion bands}.

\begin{definition}
    \label{def:cong_bands}
    Given an edge $e$ where $k$ is the maximum number of humans, a set of \emph{congestion bands}~\cite{congawarestreet} for $e$ is given by $\congbandset_e = \{\band^0,\cdots,\band^b\}$, where $\band^j = \llbracket {lb}^j, {ub}^j \rrbracket$ is an interval over the number of humans. The first band $\band^0 = \llbracket 0,0 \rrbracket$ captures an empty edge. The final band $\band^b = \llbracket {lb}^b , k \rrbracket$ is bounded by the maximum number of humans. Congestion bands do not intersect, and each number of humans fits in one band, i.e. ${lb}^{j+1} = {ub}^j + 1$.
\end{definition}

\noindent By using congestion bands, we reduce the branching factor of our planning model in Sec.~\ref{sub:planning}, improving scalability.
Further, congestion bands help when there is limited data for building duration distributions, as data is aggregated. 
At a slight abuse of notation, for a topological map $\topologicalMap$ we write $\edgedist(e,\band^j)$ to denote the duration distribution for band $\band^j$ on edge $e$.
The final congestion probabilities are denoted ${{\Pr}(\band^j\mid e@t)}$, i.e. the probability of band $\band^j$ on edge $e$ at time $t$.
We compute these by summing the values $\Pr(q\mid e@t)$ inside each congestion band:
\begin{equation}\label{eq:cong_prob}
\Pr(\band^j \mid e@t) = \sum_{q \in \band^j} \Pr(q \mid e@t).
\end{equation}

\subsection{Tour Planning under Human Congestion}\label{sub:planning}

In this subsection, we describe how to build and solve an MDP that addresses Problem~\ref{prob:tour}, i.e. tour planning under congestion.
Each navigation action in the tour MDP has a probabilistic outcome for each congestion band, and the stochastic dynamics of robot navigation are approximated using the expected duration under each congestion band.
This approximation improves the tractability of planning.
The tour MDP is similar to the single-robot MDPs used for multi-robot planning under congestion in~\cite{congawarestreet}, but adapted for a single robot in human-populated environments.

\begin{definition}
\label{def:mdp_planner_synthetic}
Given topological map $\topologicalMap$ and time bound $\timebound \in \mathbb{R}_{\geq 0}$, the tour MDP is a tuple $\mdp = \langle \mdpStates, \mdpActions, \mdpTransitionFunction, \mdpCost, \mdpGoal\rangle$, where:
\begin{itemize}
\item $\mdpStates = \vertices \times \mathbb{R}_{\geq 0} \times 2^\vertices$. States describe the robot's current node/POI, the current time, and the set of POIs already visited on the tour.
\item $\mdpInit = (\bar{v}, \bar{t}, \emptyset)$ for the first decision step, where $\bar{v}$ and $\bar{t}$ are the starting location and time for the tour, respectively. 
In subsequent decision steps, $\mdpInit$ is set to the current state.
\item $\mdpActions = \edges \,\cup\, \{\wait\}$. The robot can navigate along edges of the topological map or wait for duration $\waitdur$.
Waiting allows the robot to hold off from traversing an edge until it is less congested.
At each state the robot can navigate along edges that are outgoing from its current node.
\item The transition function $\mdpTransitionFunction$ captures the effect of robot navigation and waiting under congestion.
If a robot navigates along an edge, there is an outcome for each congestion band with non-zero probability at that time.
Upon completing a navigation action, the robot's location is updated to the destination of the edge, the destination is added to the visited set, and the time is increased by the expected edge duration under the corresponding congestion band.
Wait actions increase the time by $\waitdur$, with the location and visited set remaining the same.
Adding time to the state makes the state space infinite.
Therefore, we add a time bound $\timebound$, where actions can only be executed before $\timebound$.
Any state with a time greater then $\timebound$ becomes a dead end.
If $\timebound$ is set greater than the duration of any reasonable tour, solution quality will be unaffected.
By using time bound $\timebound$ and expected edge durations, the MDP state space becomes finite.
%
%\rewrittenSte{Moreover, to reduce the state space even more, we truncate the time to 2 decimal values to avoid floating point rounding errors.}
%
Formally, for states $s=(v,t,\visitedset)$, $s'=(v',t',\visitedset')$, and action $e$: 
\begin{equation}
\mdpTransitionFunction(s,e,s') = \begin{cases}
                                \Pr(\band^j \mid e@t) & e=(v,v'),\\
                                          & t < \timebound, \\
                                          & t' = t + \mathbb{E}[\edgedist(e,\band^j)] \text{ and} \\
                                          & \visitedset' = \visitedset \cup \{v'\} \\
                                1         & e=\wait,\\
                                          & t' = t + \waitdur,\\
                                          & t < \timebound \text{ and}\\
                                          & \visitedset' = \visitedset\\
                                0         & \text{otherwise.}
                                \end{cases} 
\end{equation}
\item The cost function $\mdpCost$ captures the expected navigation duration over all congestion bands. Formally, for state $s=(v,t,\visitedset)$ and action $e$:
\begin{equation}
    \mdpCost(s,e) = \begin{cases}
        \sum_j \Pr(\band^j \mid e@t) \cdot \mathbb{E}[\edgedist(e,\band^j)] & e \in \edges\\
        \waitdur & e = \wait.\\
    \end{cases}
\end{equation}
\item $\mdpGoal = \{(v,t,\visitedset) \in \mdpStates \mid \visitedset = \vertices \text{ and } t < \timebound\}$, i.e. the robot reaches the goal if it completes the tour within time bound $\timebound$.

\end{itemize}
\end{definition}

Though we introduce time bound $\timebound$ to constrain the state space, tour MDP $\mdp$ is still often too large to solve exactly online.
Therefore, to facilitate online planning we solve $\mdp$ using labeled real-time dynamic programming (LRTDP)~\cite{bonet2003labeled}.
LRTDP is an anytime, trial-based heuristic search algorithm that relies on an admissible heuristic $\heuristic: S \rightarrow \mathbb{R}_{\geq 0}$ to guide the search process.
For tour planning, the heuristic should be a lower bound on the remaining time to complete the tour.
We use a minimum spanning tree (MST) approach, which is a common heuristic for the travelling salesperson problem~\cite{cormen2022introduction}.
Formally, for a state $s = (v,t,\visitedset)$, we compute a new topological map $\topologicalMap_s$ that removes all nodes in $\visitedset \setminus \{v\}$, i.e. the nodes we have already visited.
Each pair of nodes $(v,v')$ in $\topologicalMap_s$ is connected, where the edge duration is the expected duration of the shortest path between $v$ and $v'$ in $\topologicalMap$, assuming no congestion.
The heuristic value $h(s)$ is then the sum of all expected edge durations on the MST for $\topologicalMap_s$.
Using the shortest path durations in $\topologicalMap_s$ allows nodes to be revisited without requiring them in the MST.

\section{EXPERIMENTS}\label{sec:experimental_evaluation}
In this section we demonstrate the efficacy of our tour planner on synthetic data and in a physical simulator.
All experiments are run in a Ubuntu 24.04 Docker image on a machine with an AMD EPYC 7513 CPU @ 2.60 GHz (max 3.65 GHz) with 32 cores and 64 threads, 1 TB of RAM and an NVIDIA A100 GPU.
An open source Python implementation of our approach can be found on Github\footnote{\url{https://github.com/convince-project/congestion-coverage-plan}}.

\subsection{Synthetic Experiments}

We begin by evaluating our planner in a synthetic simulator built from real-world human data.

\subsubsection{Experimental Setup}\label{sub:setup}

We evaluate our planner by building synthetic simulations using the \emph{ATC} shopping mall dataset~\cite{brvsvcic2013person}.
We use 50\% of the data to build a \cliffMapName{} using the implementation provided by the original authors~\cite{cliff_main,cliff-lhmp}.
To test scalability, we generate four topological maps of different sizes.
The size of each map is shown in Table~\ref{tab:Map_sizes}. %, and the largest map can be seen in Fig.~\ref{fig:atc_maps}.
%
%For the museum environment, the maps have $11$ nodes and $23$ edges; $16$ nodes and $27$ edges; $21$ nodes and $48$ edges; and $26$ nodes and $74$ edges, respectively.
%
\begin{table}[htpb]
    \centering
    \begin{tabular}{c|c}
         Nodes & Edges \\
         11 & 23 \\
         16 & 35 \\
         21 & 49 \\
         26 & 61
    \end{tabular}
    \caption{The maps created in the ATC domain.}
    \label{tab:Map_sizes}
\end{table}

All topological edges are bidirectional.
For each topological map we also consider three different sets of congestion bands, one with two bands, one with five, and one with eight.
We do this to analyse the effects of more accurate congestion models on execution-time performance and planning time.
For each set of congestion bands, we have a single congestion band for an empty edge (see Def.~\ref{def:cong_bands}), and then subdivide the remaining congestion bands equally.
These are detailed in Table~\ref{table:levels}.

\begin{table}[ht!]
    \centering
    \begin{tabular}{c|l}
       \# of bands  & Bands \\
        2 &  $\llbracket 0,0 \rrbracket$ , $\llbracket 1,max \rrbracket$ \\
        %3 &  [0,1), [1,3), [3,9999999] \\
        %4 &  [0,1), [1,3), [3,6), [6,9999999] \\ 
        5 &  $\llbracket 0,0 \rrbracket$,$\llbracket 1,2 \rrbracket$,$\llbracket 3,4 \rrbracket$,$\llbracket 5,6 \rrbracket$,$\llbracket 7,max \rrbracket$ \\
        %6 &  [0,1], [1,3], [3,5], [5,7], [7,9], [9,9999999] \\
        %7 &  [0,1], [1,3], [3,5], [5,7], [7,9], [9,12], [12,9999999] \\
        8 &  $\llbracket 0,0 \rrbracket$,$\llbracket 1,1 \rrbracket$,$\llbracket 2,2 \rrbracket$,$\llbracket 3,3 \rrbracket$,$\llbracket 4,4 \rrbracket$,$\llbracket 5,5 \rrbracket$,$\llbracket 6,6 \rrbracket$,$\llbracket 7,max \rrbracket$ \\
    \end{tabular}
    \caption{The congestion bands used in our experiments.}
    \label{table:levels}
\end{table}

To evaluate our framework, we build a synthetic simulation using the data remaining after learning the~\cliffMapName{}.
For each experimental run, we randomly sample a time from the test dataset and then replay the human movements from this point.
This human data is mapped to the edges of the topological map using the approach in Fig.~\ref{fig:edge_mapping}, where the rectangles $\rectangle_e$ for each edge $e$ are $2$m wide.
The duration for navigating an edge is the edge distance multiplied by the robot speed, plus a $10$ second penalty for each human encountered on the edge.

Our tour planner runs LRTDP online for each planning step with a convergence threshold of $0.1$, where the time bound $\timebound$ is set to $250$ seconds.
The wait duration is set to $5$ seconds based on empirical observations.
At each time step, the robot senses any humans and uses this to update the \cliffMapName{} predictions.
In particular, we use \cliffLHMPName{} to predict ten $50$ second trajectories for each observed human.
These are used to compute the congestion probabilities in~\eqref{eq:presence_prob} and~\eqref{eq:cong_prob}.
We consider two variants of our tour planner.
The first, referred to as \emph{LRTDP}, runs LRTDP until convergence or until $3000$ seconds has passed.
The second, referred to as \emph{LRTDP bounded}, runs planning for a maximum of $3$ seconds per timestep.
This is more suitable for online planning but may synthesise suboptimal behaviour.
By running LRTDP until convergence we consider the maximum performance attainable by our planner.
We compare our planner to an online Hamiltonian path solver.
Similar to our planner, this baseline re-plans after each action.
At each step, we predict the current congestion levels on each edge and assume it to be fixed.
We then use the state-of-the-art LKH 3 solver~\cite{helsgaun2017extension} to synthesise a valid Hamiltonian path for the remaining POIs.
We give LKH 3 a maximum time limit of $3000$ seconds to find the optimal solution, though it often requires less than one second on average.
This baseline allows us to evaluate the benefits of modelling stochastic crowd movement during planning.
For each topological map, set of congestion bands, and method, we run $40$ simulations sampled from the datasets and record the planning time and the total execution time.

\begin{figure}[htpb]
    \centering
        \begin{subfigure}[t]{1\linewidth}
            \centering
            \includegraphics[width=0.75\linewidth, keepaspectratio]{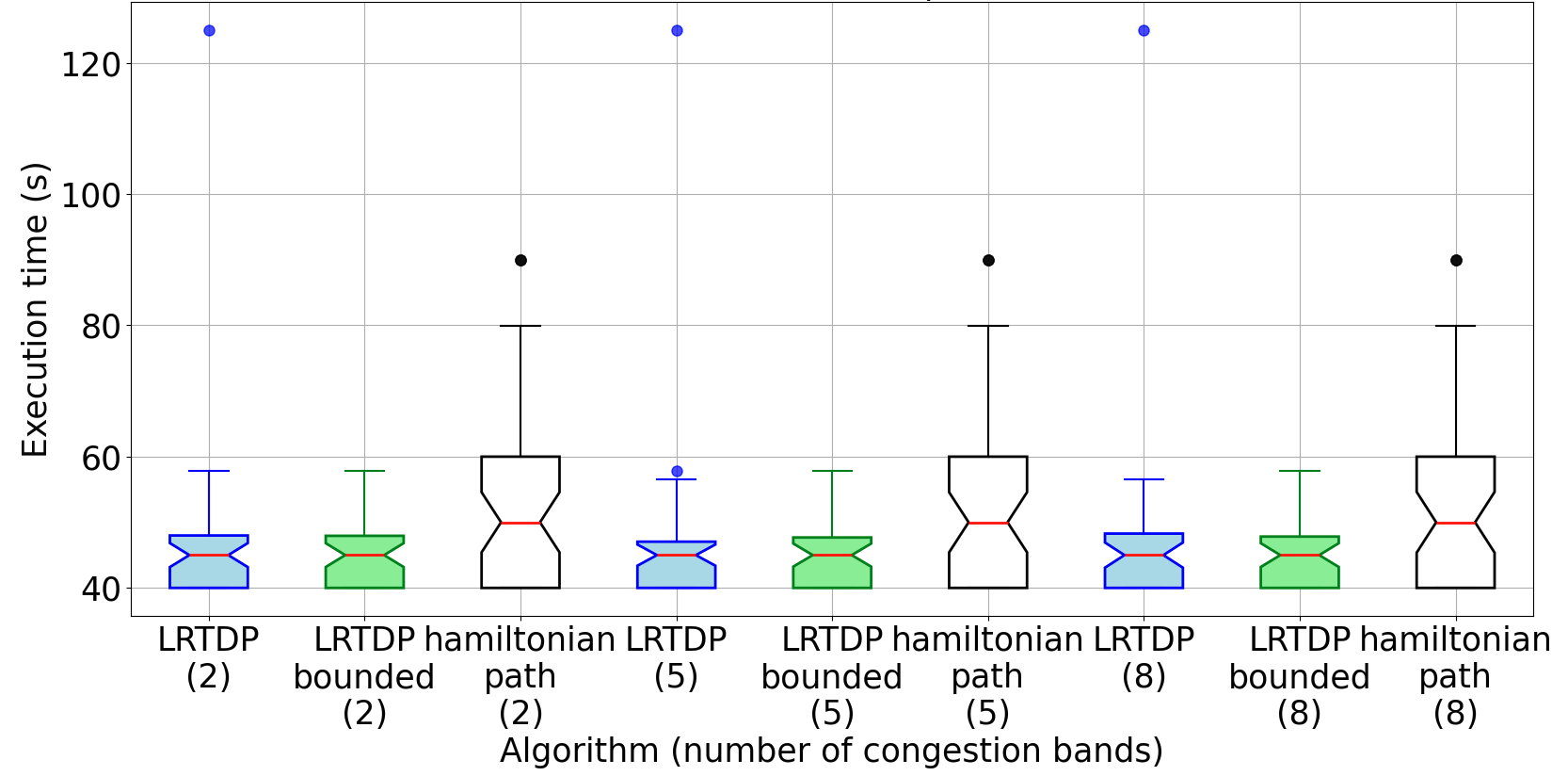}
            \caption{11 node map.}
            \label{fig:planning_time_atc_11}
    \end{subfigure}
    \begin{subfigure}[t]{1\linewidth}
            \centering
            \includegraphics[width=0.75\linewidth, keepaspectratio]{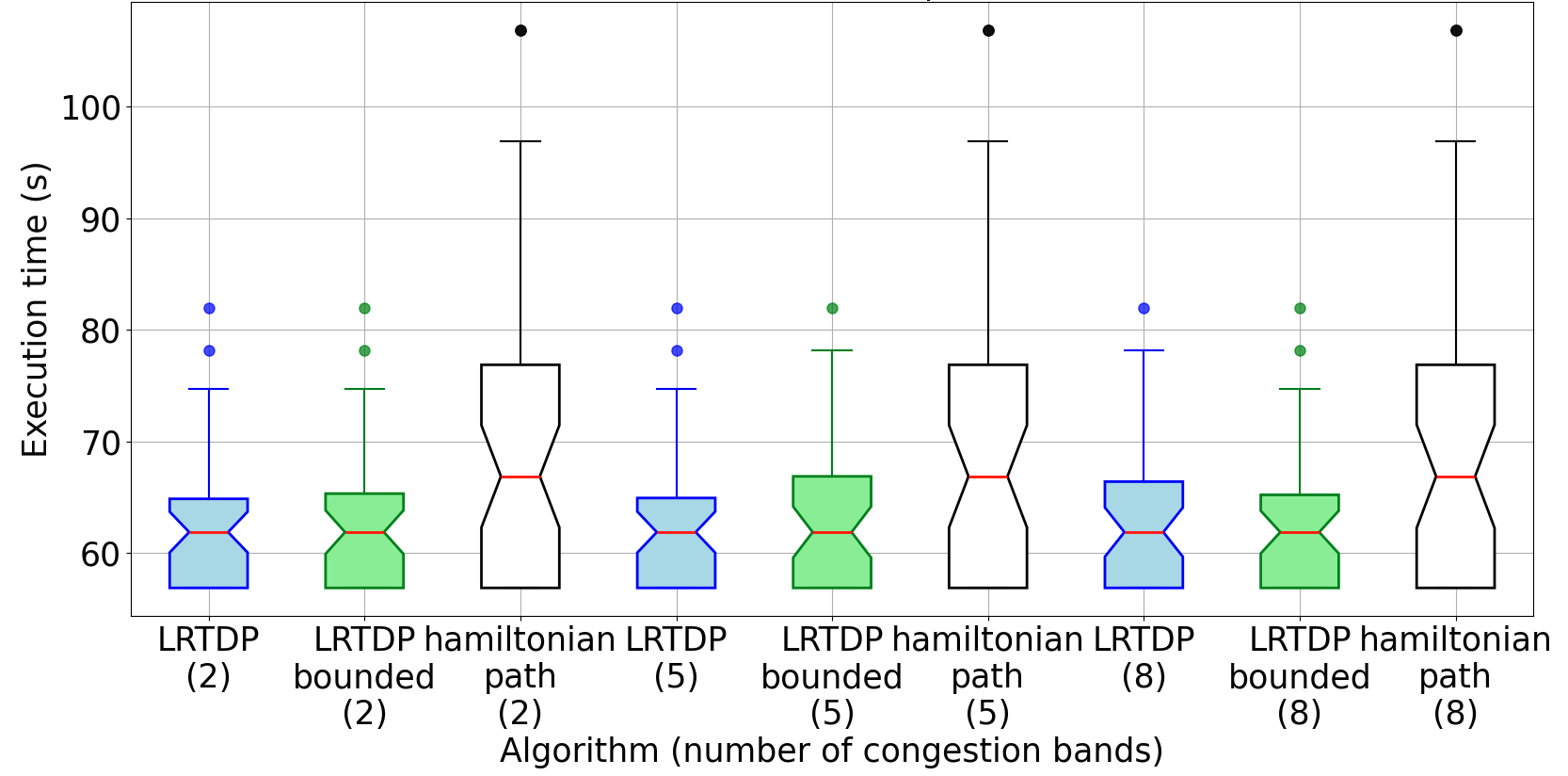}
            \caption{16 node map.}
            \label{fig:planning_time_atc_16}
    \end{subfigure}
    \begin{subfigure}[t]{1\linewidth}
            \centering
            \includegraphics[width=0.75\linewidth, keepaspectratio]{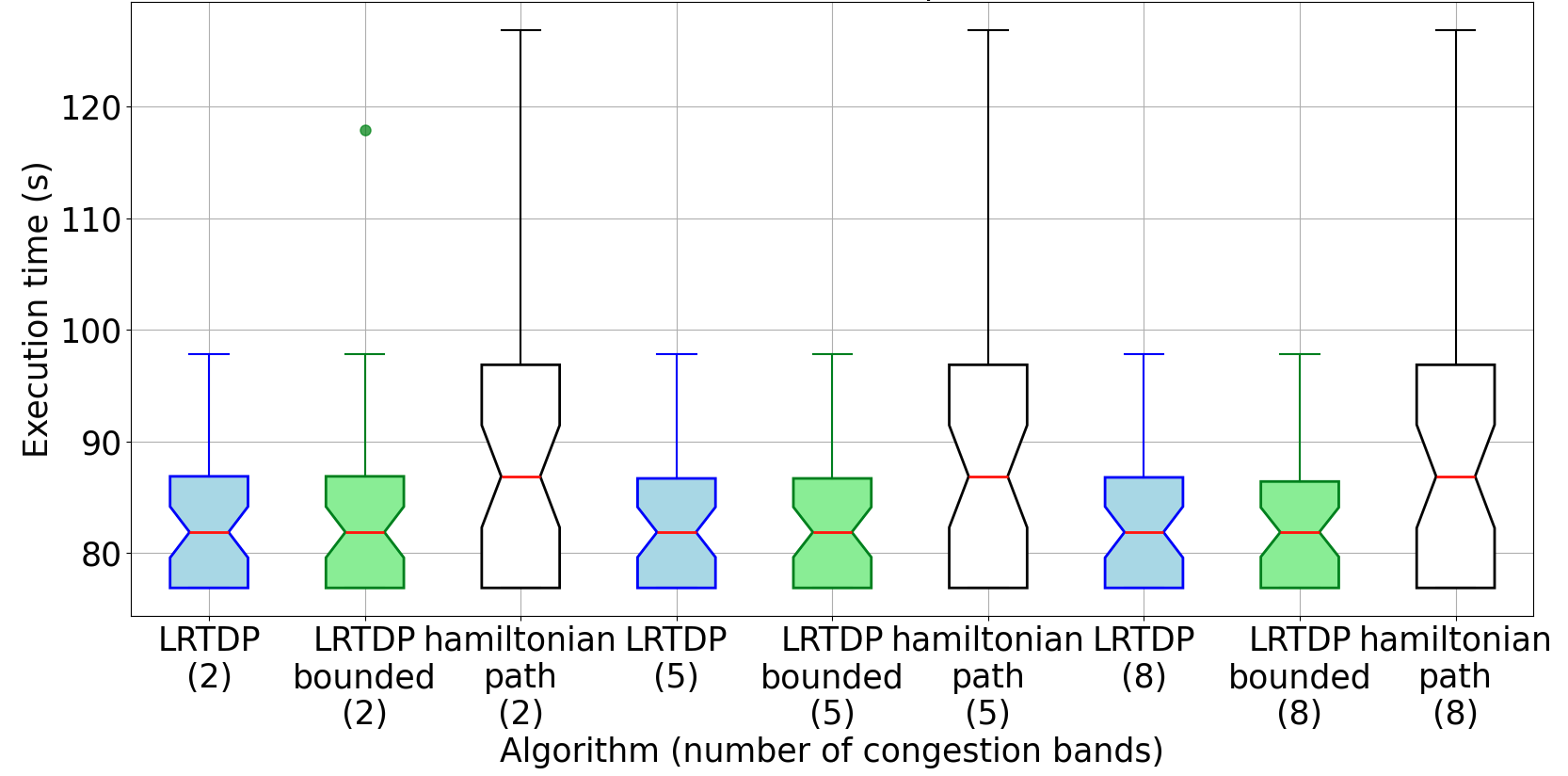}
            \caption{21 node map.}
            \label{fig:planning_time_atc_21}
    \end{subfigure}
    \begin{subfigure}[t]{1\linewidth}
            \centering
            \includegraphics[width=0.75\linewidth, keepaspectratio]{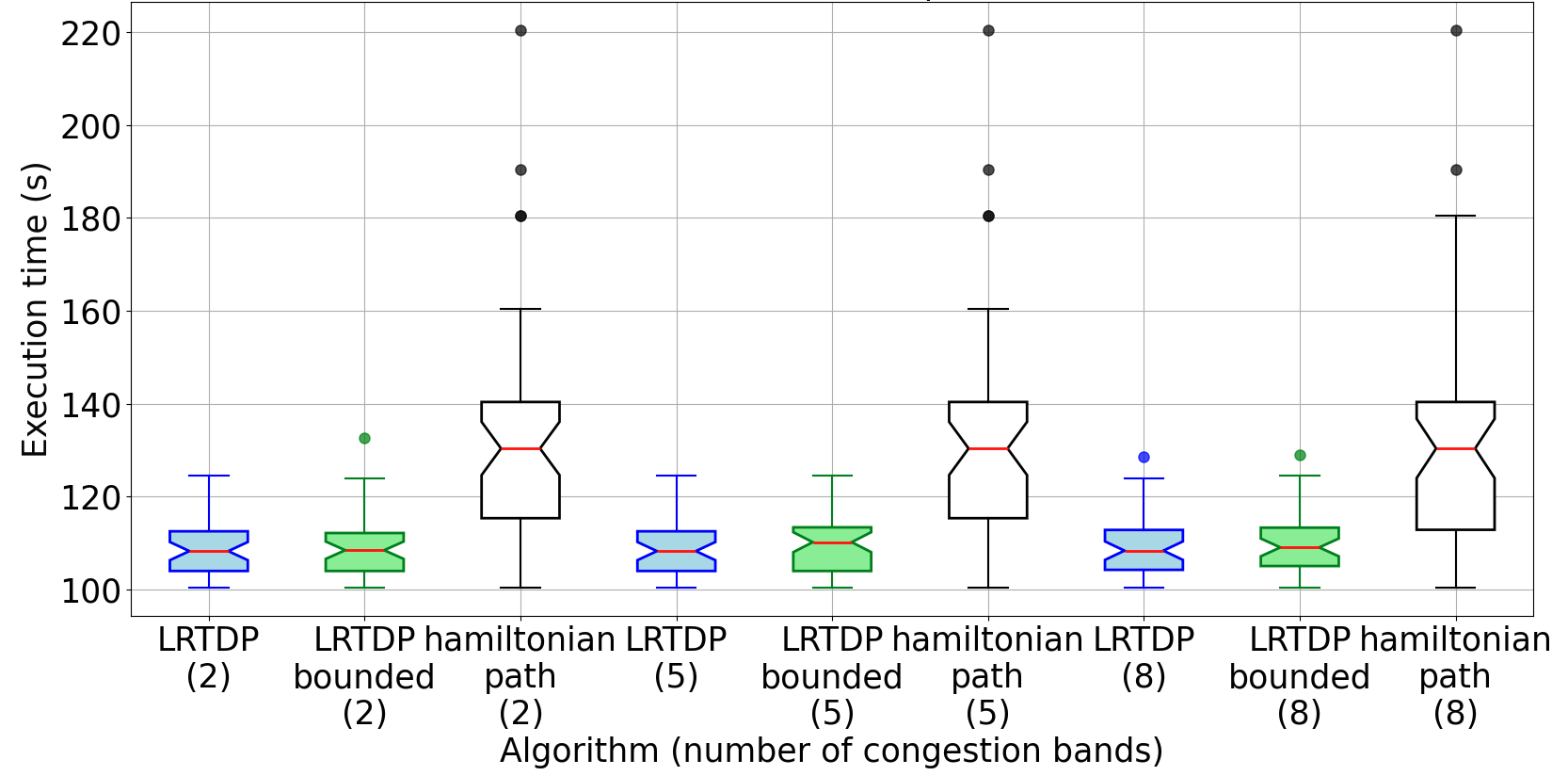}
            \caption{26 node map.}
            \label{fig:planning_time_atc_26}
    \end{subfigure}
    \caption{The execution time of the LRTDP solution (blue), the LRTDP bounded solution (green), and the hamiltonian path solution (white).}
    \label{fig:results_atc_execution_time}
\end{figure} 
% \paragraph{datasets}
% The evaluation 
% We have collected a dataset that captures the path of visitors inside the museum. This dataset was obtained from a set of LIDARs installed in the museum for several months, and by running a people detector and a tracker to extract the path followed by the visitors when the museum was open.
%
%For the map from the museum we have captured the position and the tracking of the people when the museum was open. 
% From this data we have computed the CLiFF map to predict the future position of the people. Only one map was computed from this dataset,  composed by the various points of interest that the robot has to explain inside the museum. %The maps can be seen in Figure \ref{fig:madama_maps}
\subsubsection{Results}\label{sub:results}

We present the execution time results in Fig.~\ref{fig:results_atc_execution_time}.
For each topological map and number of congestion bands, both LRTDP and LRTDP bounded complete the tour faster than the Hamiltonian path solver.
For example, on the $26$ node map, the average execution time under five congestion bands is $109.12$ seconds for LRTDP, $109.71$ seconds for LRTDP bounded, and $132.09$ seconds for the Hamiltonian path solver.
These results are statistically significant according to a one-sided Mann-Whitney U test with $p=0.05$; p-values are shown in Table~\ref{tab:p-values}.
The LRTDP approaches outperform the Hamiltonian path solver as they reason over the stochastic evolution of congestion during planning.
The Hamiltonian path solver considers a static congestion model taken at the point of planning, which quickly becomes out of date and inaccurate, producing suboptimal decisions.
In all cases, the bounded version of LRTDP produces similar execution times to the unbounded version.
This is validated in Table~\ref{tab:p-values}, where the p-values of a one-sided Mann-Whitney U test with $p=0.05$ indicates that LRTDP is never statistically significantly better than LRTDP bounded.
This shows that LRTDP is finding efficient solutions within the time bound, validating that our approach can be applied online.
Increasing the number of congestion bands has little effect on the execution-time performance of the LRTDP methods.
This suggests that a binary distinction between congested and uncongested may be an accurate enough model for planning in crowded environments.

\begin{figure}[htbp]
    \centering
    \begin{subfigure}[t]{0.49\linewidth}
    \centering
    \includegraphics[width=\linewidth, keepaspectratio]{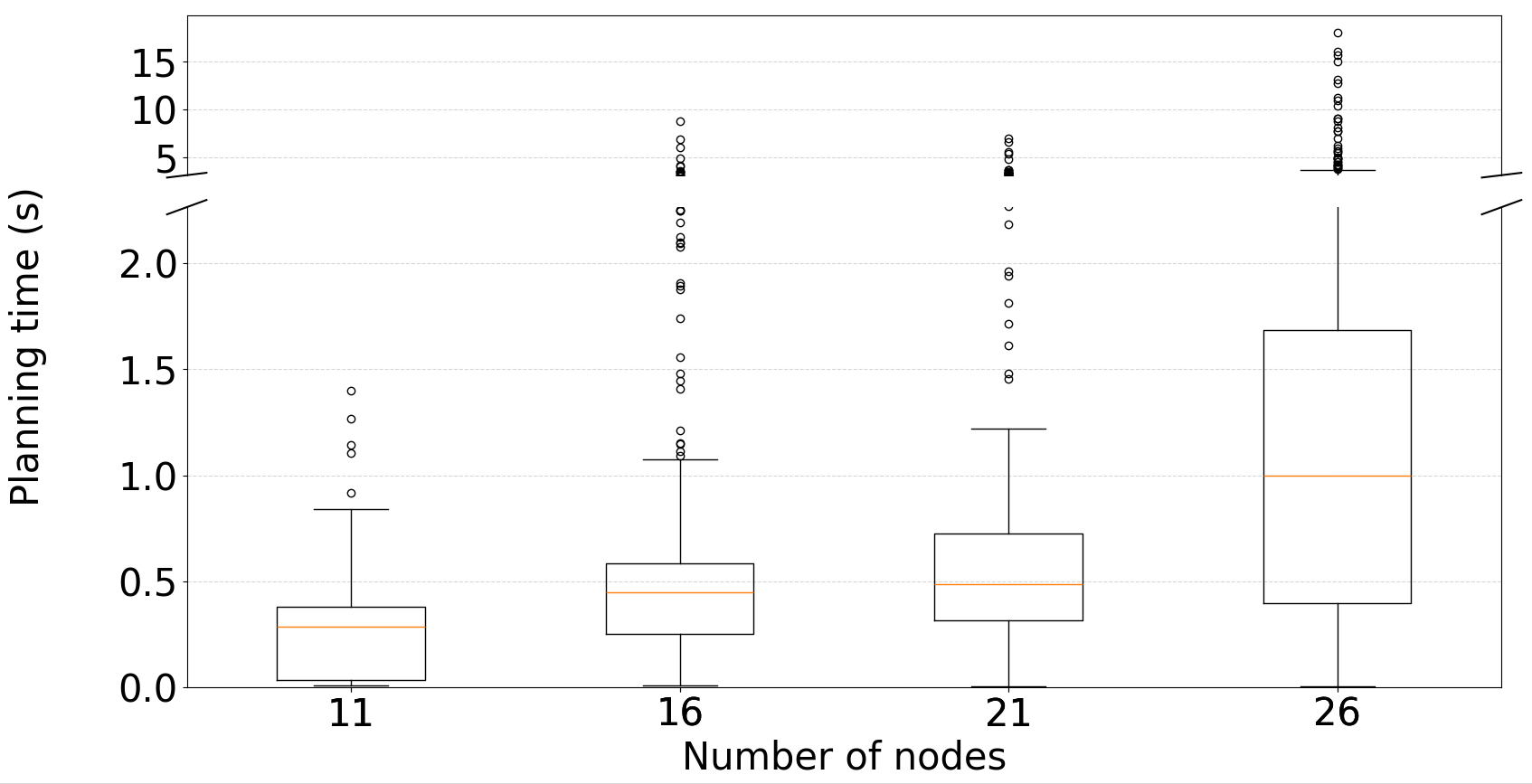}

    \caption{The planning time for the first LRTDP planning step on different map sizes. For each map we use $5$ congestion bands.}
    \label{fig:planning_times_per_vertices}
    \end{subfigure}
% \end{figure}
% \begin{figure}[htbp]
%     \centering
    \begin{subfigure}[t]{0.49\linewidth}
    \centering
    \includegraphics[width=\linewidth, keepaspectratio]{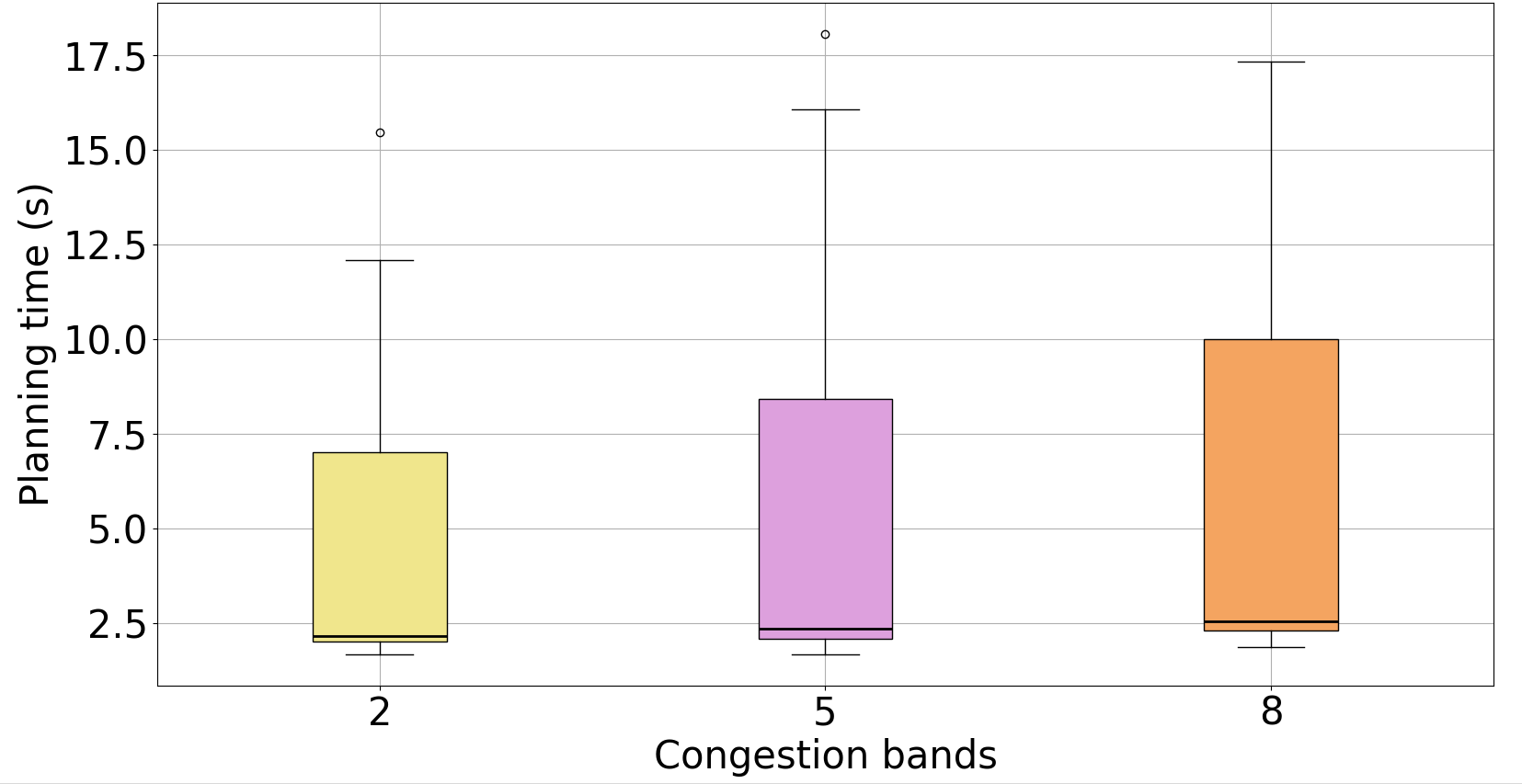}

    \caption{The planning time for the first LRTDP planning step under different numbers of congestion bands on the $26$ node map.}
    \label{fig:planning_times}
    \end{subfigure}
        \scriptsize

    \caption{LRTDP planning time results.}
    \label{fig:planning_times_all}
\end{figure}

In Fig.~\ref{fig:planning_times_per_vertices}, we show the LRTDP planning time for the first planning step on each of the four maps.
Planning time increases with the number of map nodes as both the state space and tour length increase.
However, even on the largest map the median planning time is around a second, which is suitable for online planning.
In Fig.~\ref{fig:planning_times}, we show the LRTDP planning time for the first planning step on the $26$ node map under different numbers of congestion bands.
The upper bound on planning time increases with the number of congestion bands.
This is because the MDP branching factor increases, increasing the state space.
However, in the median case the planning time is relatively consistent.
This is likely because congestion is relatively sparse in the environment and isolated to certain areas.
If the entire environment was congested at once, the branching factor increase would cause a state space explosion, as shown in the upper bound of the plots.
However, in many cases this branching factor increase only takes effect in small areas of the map, reducing the effect on planning time.

\begingroup
\renewcommand{\arraystretch}{1.25} % Default value: 1
\begin{table}[htpb]
    \centering
    \tiny
    \begin{tabular}{c|c|c|c|c}
         \multirow{2}{*}{Map} & Congestion & LRTDP $<$ & LRTDP bounded $<$ & LRTDP $<$\\
         & Bands & Hamiltonian & Hamiltonian & LRTDP Bounded\\
         \hline
         \multirow{3}{*}{26 nodes} & 2 & $\mathbf{2.67 * 10^{-9}}$ & $\mathbf{8.77 * 10^{-9}}$ & 0.43\\
        & 5 & $\mathbf{2.67 * 10^{-9}}$ & $\mathbf{1.43 * 10^{-8}}$ & 0.33\\
        & 8 & $\mathbf{6.19 * 10^{-9}}$ & $\mathbf{1.40 * 10^{-8}}$ &0.39\\
        \hline
         \multirow{3}{*}{21 nodes} & 2 & $\mathbf{8.47 * 10^{-4}}$ & $\mathbf{2.84 * 10^{-3}}$ & 0.44\\
        & 5 & $\mathbf{6.01 * 10^{-4}}$ &$\mathbf{8.49 * 10^{-4}}$ & 0.42\\
        & 8 & $\mathbf{1.09 * 10^{-3}}$ &$\mathbf{1.09 * 10^{-3}}$ & 0.58\\
        \hline
        \multirow{3}{*}{16 nodes} & 2 & $\mathbf{2.05 * 10^{-3}}$ & $\mathbf{2.95 * 10^{-3}}$ & 0.42\\
        & 5 & $\mathbf{2.31 * 10^{-3}}$ & $\mathbf{3.70 * 10^{-3}}$ & 0.45\\
        & 8 & $\mathbf{2.26 * 10^{-3}}$ & $\mathbf{1.36 * 10^{-3}}$ & 0.59\\
        \hline
        \multirow{3}{*}{11 nodes} & 2 & $\mathbf{6.25 * 10^{-3}}$  & $\mathbf{4.19 * 10^{-3}}$ & 0.54\\
        & 5 & $\mathbf{4.88 * 10^{-3}}$ & $\mathbf{3.69 * 10^{-3}}$ & 0.50\\
        & 8 & $\mathbf{5.05 * 10^{-3}}$ & $\mathbf{2.90 * 10^{-3}}$ & 0.55\\
    \end{tabular}
    \caption{The Mann-Whitney U p-values comparing LRTDP and LRTDP bounded to the Hamiltonian path solver, and p-values comparing LRTDP against LRTDP bounded. Values in bold are less than $p=0.05$ and are statistically significant.}
    \label{tab:p-values}
\end{table}
\endgroup

% values: 
% (Table \ref{tab:p-values}).

% Moving to congestion bands, we have discovered that the best number of congestion bands is 5 as can be evinced from Figure \ref{fig:execution_time_congestion_bands_atc}, where we show that the average execution time is lower than the other bands. The same result can be found in the planning time (Figure \ref{fig:planning_time_atc_congestion_bands}).

% In addition because LRTDP is an anytime planning it can be stopped in case the robot arrives at the next vertex before the optimal plan is computed. This avoids waiting for the complete computation, although it may lead to sub-optimal solutions.
%
%use the plan given to reach the next vertex and re-plan again subsequently, although this may lead to a sub-optimal solution.
% \begin{figure}[ht!]
%     \centering
%     \includegraphics[width=1\linewidth]{figures/execution time per levels atc.png}
%     \caption{Execution time per congestion band of LRTDP on the ATC dataset with the largest map}
%     \label{fig:execution_time_congestion_bands_atc}
% \end{figure}

% \begin{figure}[ht!]
%     \centering
%     \includegraphics[width=1\linewidth]{figures/MUSEUM_RESULTS.png}
%     \caption{Execution time compared to the hamiltonian path solution on the museum dataset.}
%     \label{fig:results_museum_execution_time}
% \end{figure} 

\subsection{Gazebo Experiments}

\begin{figure}[t]
    \centering
        \begin{subfigure}[b]{0.44\linewidth}
            \centering
            \includegraphics[width=\linewidth, keepaspectratio]{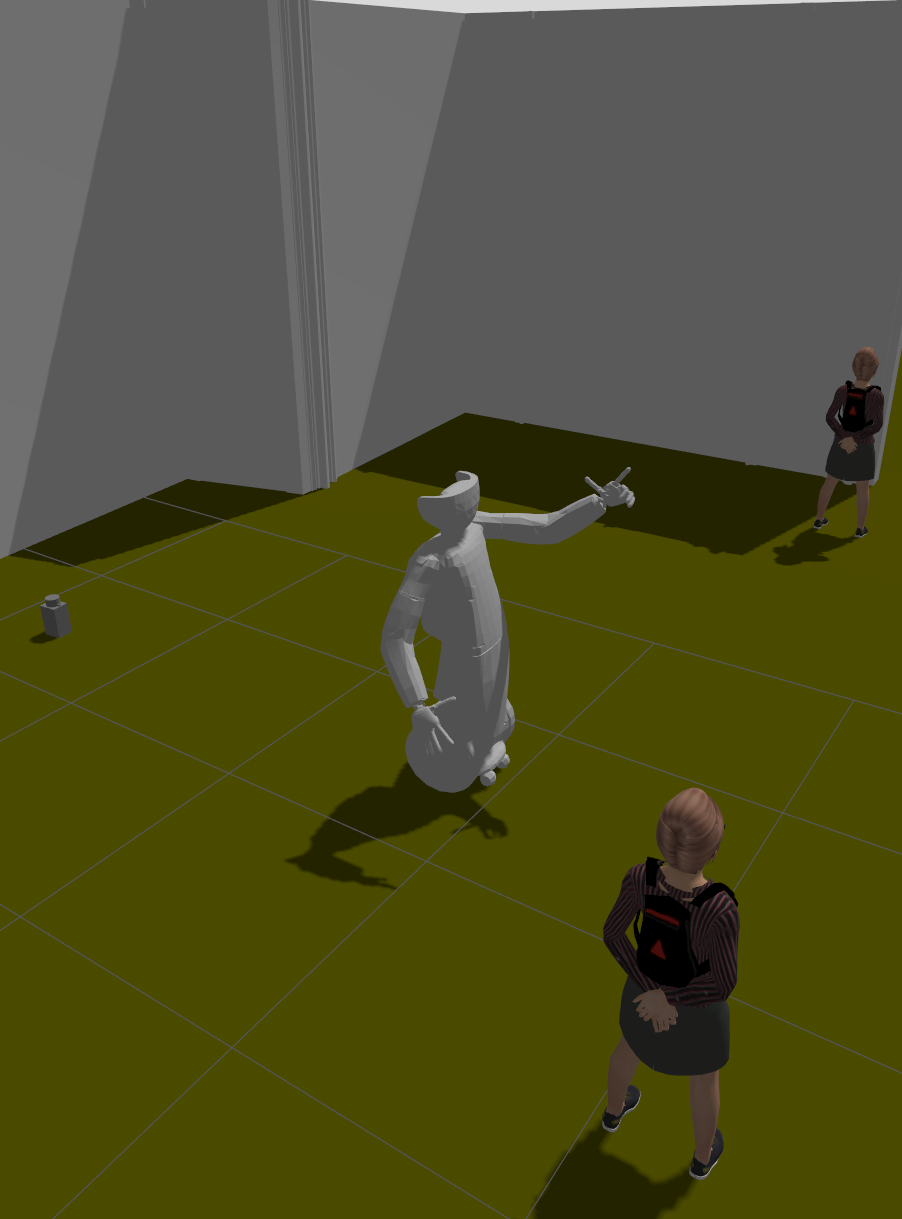}
            \caption{Gazebo simulation.}
            \label{fig:gazebo}
    \end{subfigure}
            \begin{subfigure}[b]{0.495\linewidth}
            \centering
            \includegraphics[width=\linewidth, keepaspectratio]{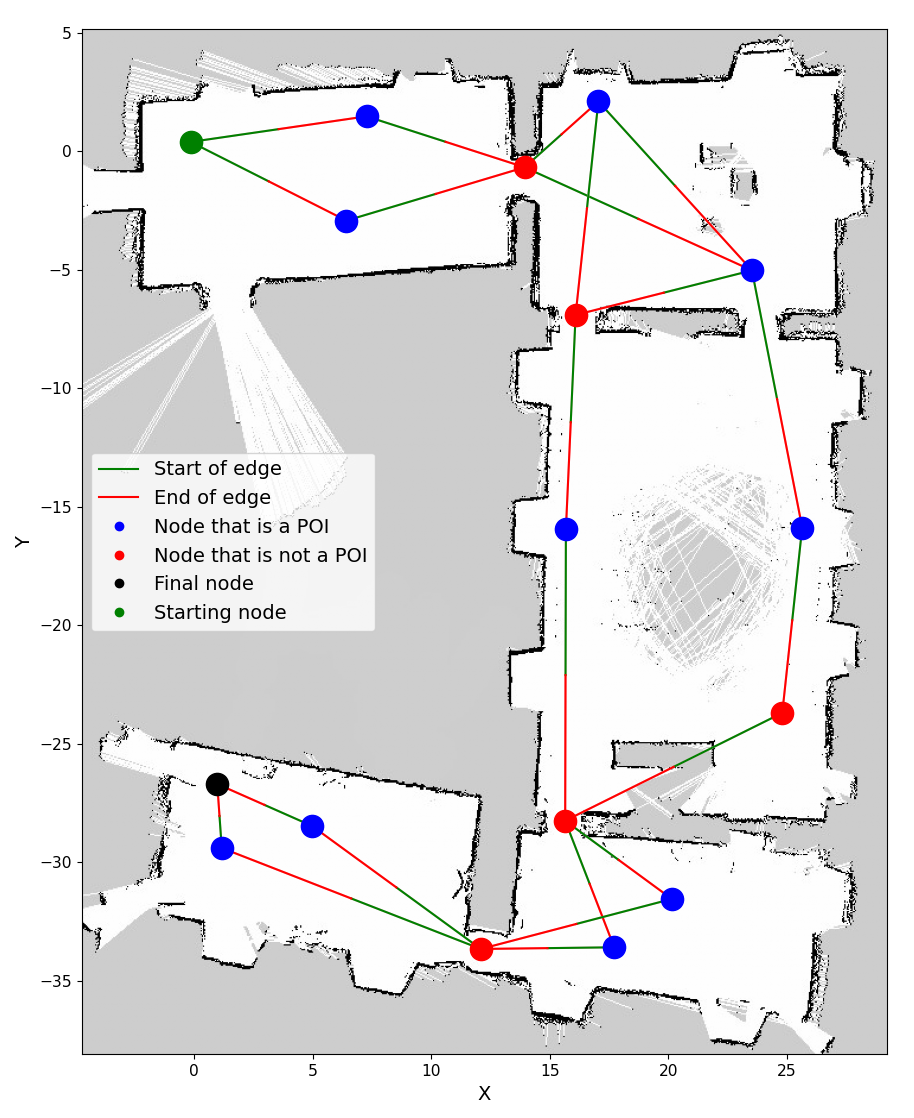}
            \caption{Topological map.}
            \label{fig:gazebo_top_map}
    \end{subfigure}
    \caption{The Gazebo simulation of the Palazzo Madama and the corresponding topological map.}
    \label{fig:gazebo_exp}
\end{figure}

Next, we evaluate our planner in a Gazebo simulation of the Palazzo Madama in Turin (see Fig.~\ref{fig:gazebo_exp}).
Here, a mobile robot provides guided tours through the museum's exhibits.

\subsubsection{Experimental Setup}
Humans are simulated by sampling trajectories from real data collected in the Palazzo Madama.
The topological map has $17$ nodes, where $10$ are POIs (see Fig.~\ref{fig:gazebo_top_map}).
Supporting this domain requires minor adjustments to our planner.
First, upon reaching a POI the robot must execute an action explaining it. 
This must be modelled in the MDP and LRTDP heuristic to capture the passage of time, which is crucial for congestion modelling.
Next, since the Palazzo Madama is small with few routing options, we allow the robot to explain a POI from multiple locations, allowing it to choose less congested areas.
The visited set $\visitedset$ is updated in the MDP once a POI is explained at one of its possible locations.

We compare our approach against a baseline which finds the shortest route through the POIs without considering congestion.
For each method, we run $30$ tours and evaluate the total tour duration and the number of navigation failures per tour.
Navigation failures are a proxy for how well the robot avoids congestion, as navigation is more likely to fail in congested areas.

\iffalse
The algorithm was also tested in a simulated environment, which is a Gazebo reproduction of Palazzo Madama in Turin. Given that  the path was forced to be sequential, we decided to adapt the algorithm as Sec. \ref{sub:alternative_pois}.
\subsection{Setup}

The tests were performed on the same machine as Sec. \ref{sec:experimental_evaluation}.
%
To assess the impact of the congestion-aware planner, we analysed the robot's performance across 60 full museum tours: 30 with the planner enabled (\textit{w/ LRTDP}) and 30 with the planner disabled (\textit{w/o LRTDP}) where the robot choose a default POI.
%
Evaluation was carried out in a simulated environment replicating the museum
%and executing the same code used on the real system 
where visitors are simulated  according to trajectories extracted from a collected dataset.
The map used was composed by 17 vertices (of which 10 were POIs) and ... edges
%
The evaluation metrics gathered were total tour duration, the number of navigation recoveries, and the number of navigation aborts.

\begin{figure}
    \centering
    \includegraphics[width=0.5\linewidth]{figures/replan_gazebo.png}
    \caption{The gazebo simulation}
    \label{fig:gazebo_simulation}
\end{figure}
\begin{figure}
    \centering
    \includegraphics[width=0.5\linewidth]{figures/replan_pois_no_names.png}
    \caption{The map used for the gazebo simulation}
    \label{fig:gazebo_map}
\end{figure}
\fi

\subsubsection{Results}\label{sec:gazebo_planner_evaluation}
\begin{figure}[t]
    \centering
        \begin{subfigure}[t]{0.49\linewidth}
            \centering
            \includegraphics[width=\linewidth, keepaspectratio]{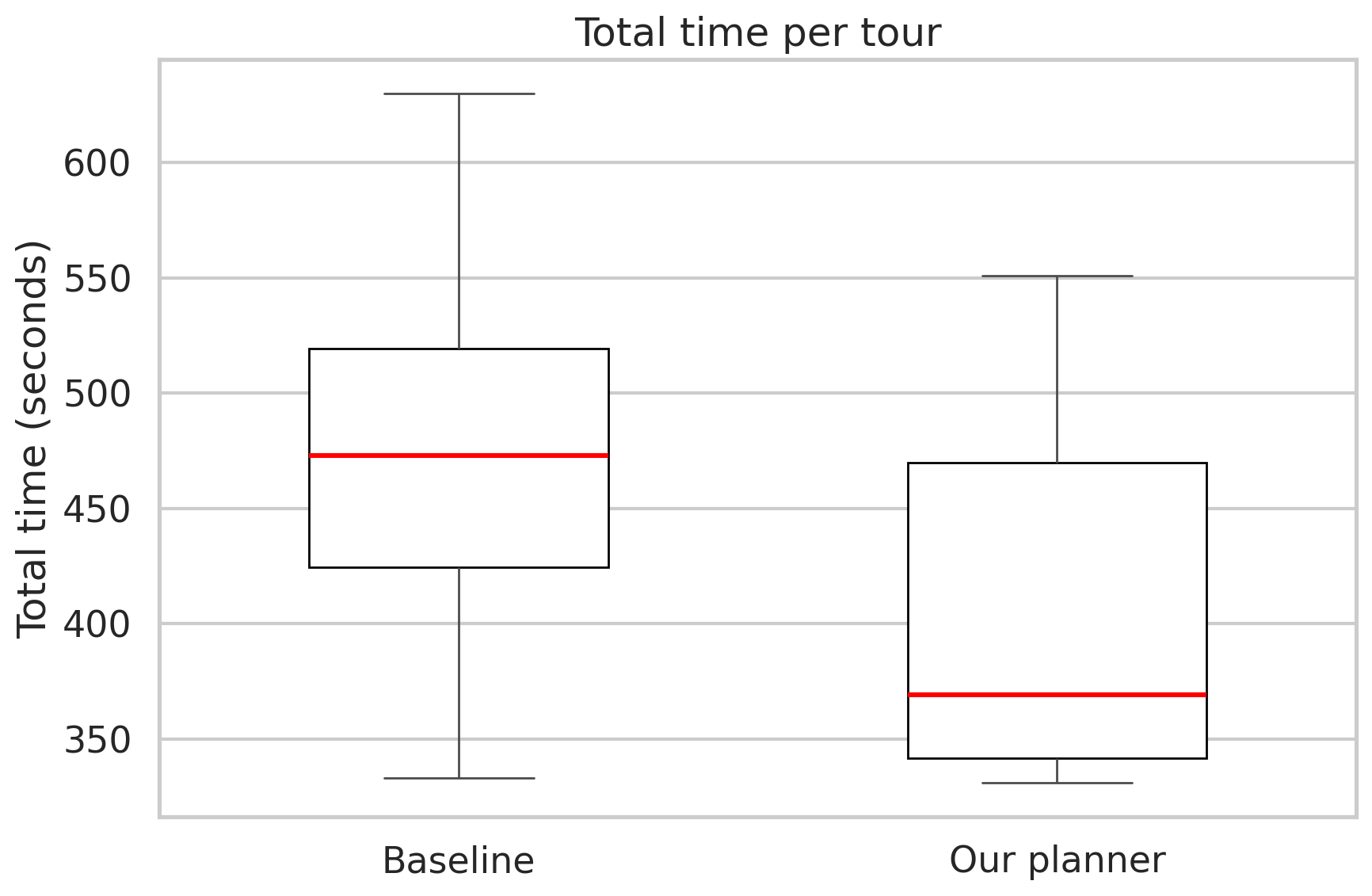}
            \caption{Tour completion time.}
            \label{fig:replan_time_per_tour}
    \end{subfigure}
            \begin{subfigure}[t]{0.49\linewidth}
            \centering
            \includegraphics[width=\linewidth, keepaspectratio]{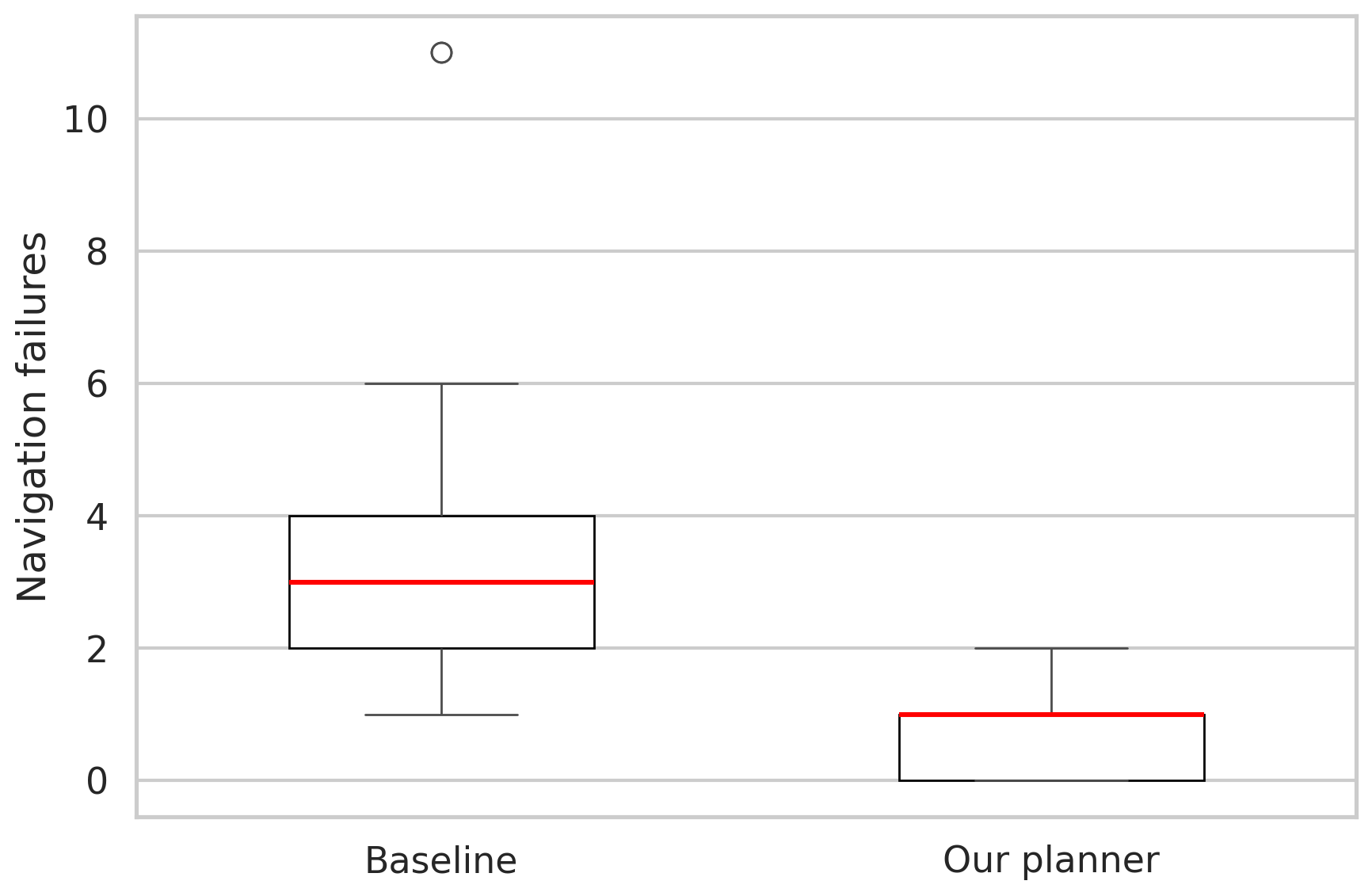}
            \caption{Navigation failures per tour.}
            \label{fig:replan_aborts_per_tour}
    \end{subfigure}
    \caption{Gazebo simulation results.}
    \scriptsize
    \label{fig:uc3-replan_time_per_tour}
\end{figure}

We present the Gazebo results in Fig.~\ref{fig:uc3-replan_time_per_tour}.
Our planner outperforms the non-LRTDP baseline in terms of both tour duration and navigation failures.
These results are statistically significant according to a one-sided Mann-Whitney U test with $p = 0.05$ (see Table~\ref{table:uc3-replan-stats}).
Rather than subscribe to a fixed route, our planner reasons over the evolution of congestion and re-routes the robot when necessary.
This reduces the effects of congestion, reducing navigation time and failures, as the robot travels through less congested parts of the environment.

In summary, our planner efficiently routes the robot by predicting future human congestion, and reacting to it as it occurs.
Our planner is also scalable enough for online planning, where the planning time can be bounded with only a small effect on execution-time performance.

%The graphs in Figure~\ref{fig:uc3-replan_time_per_tour} show an advantage for LRTDP on all three metrics.
%
%The difference is particularly noticeable in the number of recoveries and aborts, resulting in smoother trajectories with fewer interruptions and fewer visually awkward stops to recompute paths around visitors.
%
%For these metrics, not only does the average value decrease significantly, but the distribution of events also becomes more concentrated.
%
%Overall, the navigation system becomes substantially more stable.

% \begin{table}[b]
% \centering
% \small
% \begin{tabular}{|l|c|c|}
% \hline
% Metric & w/ LRTDP & w/o LRTDP \\
% \hline
% Average total time (s)      & 405.43 & 466.16 \\
% \hline
% Average recoveries          & 13.40  & 74.23 \\
% \hline
% Average aborts              & 0.77   & 3.361 \\
% \hline
% \end{tabular}
% \caption{Statistics on 30 simulated tours with and without congestion replanning}
% \label{table:uc3-replan-results}
% \end{table}

\begin{table}[ht]
\centering
\small
\begin{tabular}{|l|c|}
\hline
Metric & p-value \\
\hline
Tour duration      & $\mathbf{1.5 * 10^{-3}}$ \\
\hline
Navigation failures              & $\mathbf{3.04 * 10^{-9}}$ \\
\hline
\end{tabular}
\caption{The Mann-Whitney U p-values comparing our planner against the baseline in the Palazzo Madama domain. Values in bold are less than $p=0.05$ and are statistically significant.}
\label{table:uc3-replan-stats}
\end{table}

\section{CONCLUSION}
\label{sec:conclusion}

In this paper, we presented a probabilistic congestion-aware tour planner for crowded environments.
We use a CLiFF map trained on real-world human data to predict human trajectories, and then use these predictions to build and solve an MDP that explicitly captures the effects of crowds on robot navigation.
Our planner operates online and is reactive to humans entering or leaving the environment.
Empirically, our approach synthesizes quicker tours than classical hamiltonian path solutions adapted to consider congestion.
In future work, we will investigate how to scale to larger tours and environments, and consider how the human prediction horizon can be improved during planning.

%We have presented a framework to plan a tour over occupancy, that on average outperform a standard hamiltonian path solution by reducing the travelling time of more than 50 seconds compared to the better hamiltonian path solution. In addition we have demonstrated also that the planning time is on average very low even when the number of nodes of the graph increases, although this is not necessary because a tour can be done with less than 15 nodes, as evinced from the museum tour.

% \section{Acknowledgments}
% Acknowledgments removed for double blind reasons.
% This work was funded by the European Union under the Horizon Europe grant 101070227 (CONVINCE). Charlie Street and Masoumeh Mansouri are UK participants in Horizon Europe Project CONVINCE and supported by UKRI grant number 10042096.
% \clearpage
% \addtolength{\textheight}{-12cm} 
%\bibliographystyle{ACM-Reference-Format} 
\bibliographystyle{ieeetr}
\bibliography{bibliography}
%\printbibliography
% \bibliographystyle{aaai2026}
% \bibliography{bibliography}

\end{document}